\documentclass{article}

 \usepackage[dblblindworkshop, final]{neurips_2025}
\workshoptitle{What Can('t) Transformers Do?}

\usepackage[utf8]{inputenc} % allow utf-8 input
\usepackage[T1]{fontenc}    % use 8-bit T1 fonts
\usepackage{hyperref}       % hyperlinks
\usepackage{url}            % simple URL typesetting
\usepackage{booktabs}       % professional-quality tables
\usepackage{amsfonts}       % blackboard math symbols
\usepackage{nicefrac}       % compact symbols for 1/2, etc.
\usepackage{microtype}      % microtypography
\usepackage{xcolor}         % colors
\usepackage{amsmath}
\usepackage{graphicx}
\title{An empirical study on the limitation of \\ Transformers in program trace generation}
\author{%
Simeng Sun \\
\texttt{simengs@nvidia.com}
}

\begin{document}

\maketitle

\begin{abstract}
We study Transformers on the task \emph{program trace generation} (PTG), where models produce step-by-step execution traces for synthetic programs. Unlike existing algorithmic problems, PTG externalizes reasoning through long traces where each step is trivial. We train small Transformers with diverse modifications, including alternative position encodings, softmax replacements, hybrid model, and short convolutions. While these models achieve strong in-distribution accuracy, they exhibit systematic failures when generalizing to various factors (e.g., program length, trace steps), though some designs significantly improve generalization.  
\end{abstract}
\section{Introduction}
\vspace{-0.3em}
We study Transformers on the task \emph{program trace generation} (PTG): given a synthetic program and inputs, models must output the complete step-by-step execution trace. This setup offloads the \emph{internal} reasoning (e.g., sorting, tracking permutations) onto long, structured traces/scratchpads that explicitly store the states of intermediate steps. 
Following prior work~\citep{l0-bench}, we generate simple, executable programs with additional constraints to reduce difficulty at each step. 
Complex execution thus emerges from composition of simple steps, which enable evaluation of models' contextual flexibility at executing arbitrary programs, contrasting with fixed-procedure tasks (e.g., large number multiplication) whose complexity is tightly coupled with input size. 
With synthetic data, we train small Transformers ($\sim$154M parameters) with diverse modifications. Evaluation is limited below training sequence length to avoid confounding of length extrapolation. While all models achieve strong in-distribution accuracy, they struggle to reliably generalize to various factors. Experimental results reveal substantial impacts of architectural choices on out-of-distribution performance, with NAPE (mixing NoPE and ALiBi heads) on average outperforming other modifications. Our results expose limitations in existing Transformers at executing programs step by step, an ability crucial to reasoning tasks and for precise instruction-following in large language models.

\section{Program trace generation}
\vspace{-0.3em}
The synthetic programs are standalone functions varying code segments among assignment, binary operations,\footnote{We further restrict arithmetic ops to be only $x+1$ and $x-1$, comparison ops to only $==$ and $!=$.} if block, while block, list append or pop. We generate these examples following the methodology of L0-Bench~\citep{l0-bench}. More details are provided in Appendix~\ref{sec:append-data}. With constraints on the simplicity at each step, execution of complex programs becomes sequentially composing basic operations, and the trace steps grow at least linearly with the input size. Therefore, compared to tasks requiring direct production of final answers, we focus on the case where models leverage computations that change with problem complexity. The research question thus shifts from ``Can Transformers execute a fixed program \textit{internally} given unseen/longer input?'' to ``Can Transformers execute simple operations consistently over long sequence given arbitrary programs?'' 
\section{Experimental setup}
\vspace{-0.2em}
\paragraph{Data \& Evaluation metrics.} 
Training and evaluation data are generated while controlling four factors: program length, trace steps, number of variables, and input size. Training data contains programs of 5 to 40 lines, 5 to 103 trace steps, maximum 10 variables, and input list of 5 to 10 integers. Short examples are concatenated together, while longer examples do not exceed 4096 tokens. For out-of-distribution evaluation, we vary each factor independently while keeping others fixed: we test on \emph{longer programs}, \emph{longer traces}, \emph{more variables}, and \emph{longer lists} (i.e., larger input size). Motivation of studying these factors is in Appendix. We report whole-trace accuracy: a trace is correct only if every step matches the ground truth exactly. More details are provided in Appendix~\ref{sec:append-data}.
\vspace{-0.2em}
\paragraph{Model \& Training.} We train small Transformers of size $\sim$154M parameters on a mixture of short and medium length programs. In addition to standard RoPE-based Transformers, we also experiment with other position encodings (NoPE, ALiBi, NaPE, Fox, PaTH), softmax replacements (STB, $\alpha$-entmax), short 1d-convolution (Canon) and hybrid model (SWAN). Description of these methods are provided in Appendix~\ref{sec:mdl-desc}. We perform fixed-budget model comparison with 20k training steps, which is sufficient for the baseline (NoPE) to achieve near perfect in-distribution performance. We focus on OOD performance to explore the compatibility of various inductive biases with PTG. All models are trained with 4k-token sequences and vocabulary size of 176. We report the average and standard error of each model across 10 random seeds. More details are provided in Appendix~\ref{sec:append-exp}.
% \vspace{-0.4em}

\section{Results}

\begin{table}[h!]
    \small
    \centering
    \vspace{-1em}
    \caption{Whole-trace accuracy (\%) across test conditions. ID: in-distribution test splits.}
    \scalebox{0.97}{
    \begin{tabular}{@{}lcc|ccccc@{}}
    \toprule
& \textbf{\begin{tabular}[c]{@{}c@{}}ID \\ (short)\end{tabular}} & \textbf{\begin{tabular}[c]{@{}c@{}}ID\\ (mid) \end{tabular}} & \textbf{\begin{tabular}[c]{@{}c@{}}Longer\\ lists\end{tabular}} & \textbf{\begin{tabular}[c]{@{}c@{}}More\\ variables\end{tabular}} & \textbf{\begin{tabular}[c]{@{}c@{}}Longer\\  traces - 1\end{tabular}} & \textbf{\begin{tabular}[c]{@{}c@{}}Longer \\  traces - 2\end{tabular}} & \textbf{\begin{tabular}[c]{@{}c@{}}Longer \\  programs\end{tabular}} \\ \midrule
NoPE   & $99.8_{\pm0.2}$  & $99.0_{\pm0.4}$& $90.3_{\pm2.5}$ &$94.1_{\pm2.0}$  & $91.6_{\pm2.7}$   &   $14.2_{\pm2.5}$   & $46.5_{\pm2.2}$    \\
RoPE   & $98.0_{\pm1.1}$  & $92.6_{\pm3.1}$& $65.5_{\pm4.2}$ &  $92.9_{\pm3.5}$  & $78.5_{\pm5.2}$   & $3.9_{\pm1.0}$    &  $40.8_{\pm3.3}$    \\
ALiBi  & $99.4_{\pm0.4}$  & $96.8_{\pm1.9}$& $86.2_{\pm3.5}$ & $92.1_{\pm1.7}$  &$93.1_{\pm1.5}$   &   $39.5_{\pm4.7}$    & $50.8_{\pm2.2}$    \\
NaPE   & $99.7_{\pm0.3}$  & $98.3_{\pm1.6}$& $93.4_{\pm1.6}$ & $98.4_{\pm0.9}$  &$97.9_{\pm1.3}$   &  $93.5_{\pm1.4}$   &  $57.4_{\pm2.1}$    \\
    Fox    & $99.2_{\pm0.4}$  & $95.8_{\pm2.1}$& $90.0_{\pm3.3}$ & $95.8_{\pm1.8}$  &$77.9_{\pm1.5}$   &   $19.2_{\pm1.7}$   & $42.9_{\pm1.1}$    \\
    PaTH & 100.0$_{\pm 0.0}$	&  99.2$_{\pm 0.2}$	& 89.7$_{\pm 3.8}$	& 90.7$_{\pm 3.1}$	& 88.8$_{\pm 2.4}$	& 30.9$_{\pm 5.3}$	& 52.0$_{\pm 4.5}$ \\
SWAN   & $97.9_{\pm1.1}$  & $85.3_{\pm3.6}$& $29.8_{\pm3.9}$ & $80.5_{\pm2.6}$  &$49.5_{\pm3.9}$   &  $0.5_{\pm0.2}$    &  $21.9_{\pm2.5}$    \\
    NoPE+Canon & 1$00.0_{\pm0.0}$ & 1$00.0_{\pm0.0}$& $58.0_{\pm8.1}$ & $98.4_{\pm0.5}$  &$95.3_{\pm1.4}$   &   $35.3_{\pm6.9}$   & $64.8_{\pm2.4}$    \\
    STB & 99.1$_{\pm 0.5}$ & 82.2$_{\pm 4.4}$ & 81.9$_{\pm 4.4}$ & 50.7$_{\pm 3.8}$ & 66.4$_{\pm 2.7}$ & 11.2$_{\pm 1.5}$ & 21.6$_{\pm 3.7}$ \\
NoPE+$\alpha$-Entmax & 100.0$_{\pm 0.0}$ & 99.1$_{\pm 0.1}$ & 41.6$_{\pm 2.2}$ & 98.2$_{\pm 0.3}$ & 98.0$_{\pm 0.4}$ & 27.1$_{\pm 1.8}$ & 49.5$_{\pm 0.7}$ \\
    \bottomrule
    \end{tabular} }
    \label{tab:main-table}
    
    \end{table}

Table~\ref{tab:main-table} shows whole-trace accuracy across architectures and test conditions. All models achieve strong in-distribution accuracy but vary significantly in generalization performance. Reliably generating long traces remains challenging across models with large degradation in whole-trace accuracy when increasing output trace steps (\emph{longer traces - 1} vs. \emph{- 2}). Models also struggle with more complex/longer instructions in the context: none reaches above 70\% accuracy. Dealing with larger entity size (\emph{longer lists}), a factor often studied by existing work, is heavily impacted by positional encoding, with standard RoPE falling behind recent variants, such as Fox and PaTH. Despite its simplicity, NaPE is surprisingly powerful at generating longer traces, being the only variant achieving over 90\% accuracy while others are below 40\%. Finally, Canon layers improve performance on longer programs (64.8\% for NoPE+Canon) but lead to degradation on longer lists (90.3\% vs. 58.0\%). 
\section{Discussion}

We empirically studied Transformers on program trace generation, a new task requiring precise rule-following over long sequences of simple computational steps. We compare multiple modifications to standard Transformers, including alternative position encoding methods, softmax replacements and novel layer designs. Experimental results show that while these models achieve near-perfect in-distribution performance, they struggle to maintain global accuracy, especially for longer programs and longer traces unseen during training. Architectural choices also significantly impact performance, with NaPE outperforming others. In the future, we plan to extend our work to latest architectures potentially including non-Transformers and MoE models. Finally, we also plan to explore the transduction setting, which may require additional modeling capabilities atop contextual flexibility.

\bibliographystyle{acl_natbib.bst}
\bibliography{ref}

\appendix
\section{Program Trace Generation} \label{sec:append-data}

\begin{figure}[h]
    \centering
    \caption{An example of program trace generationt ask.}
    \includegraphics[width=0.35\linewidth]{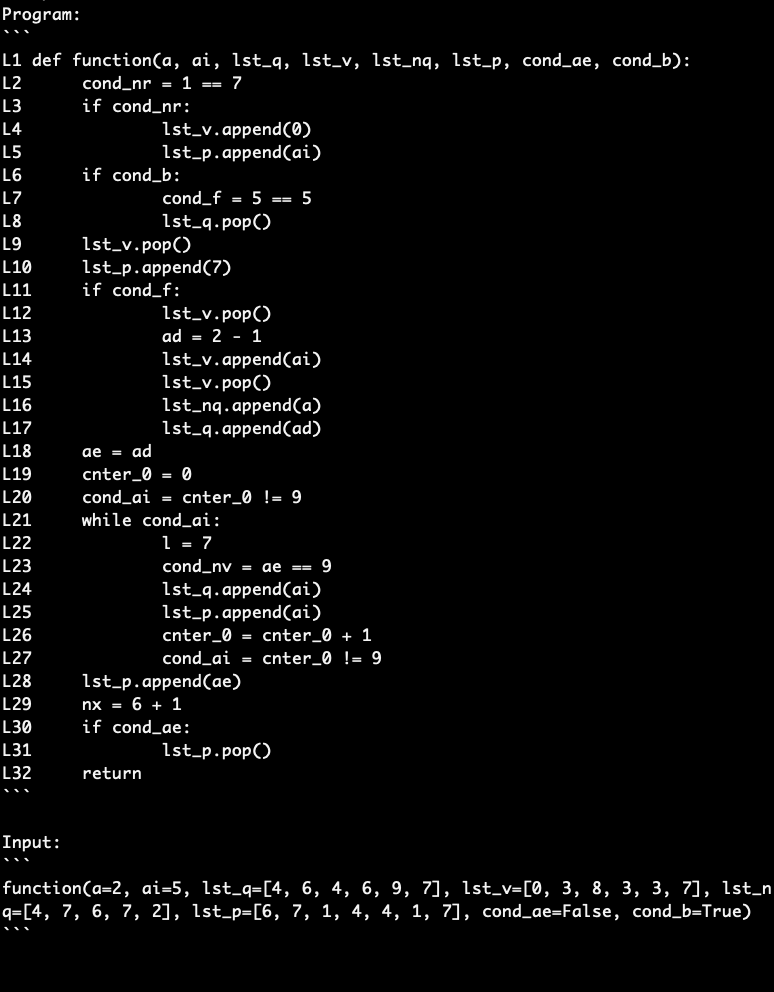}\includegraphics[width=0.35\linewidth]{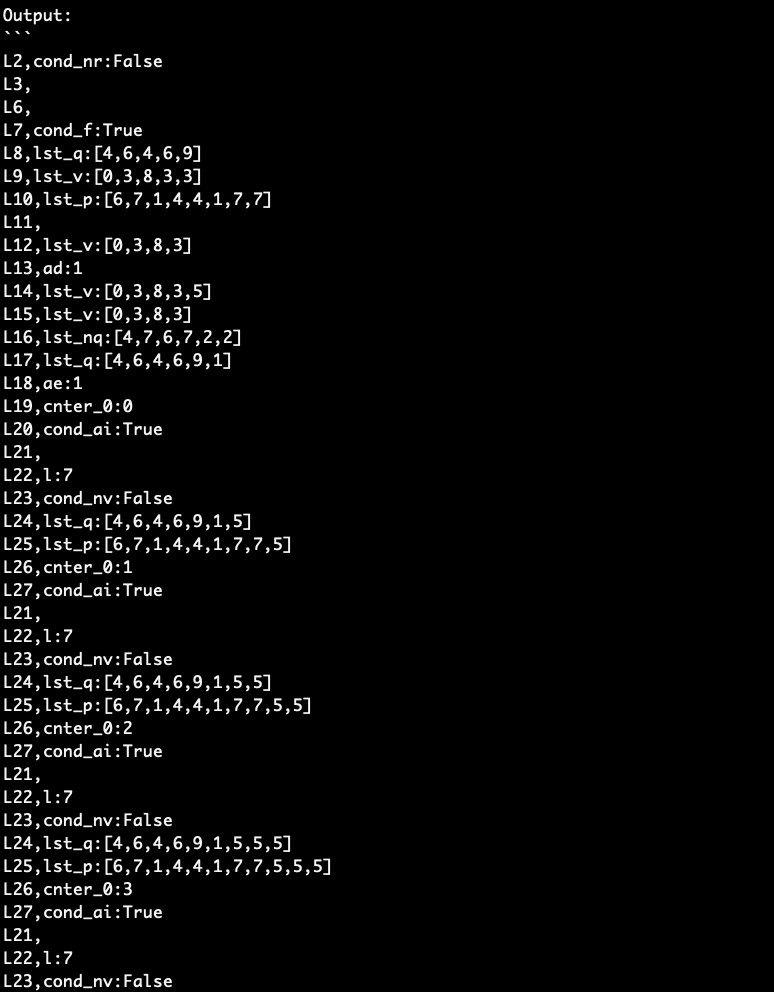}\includegraphics[width=0.35\linewidth]{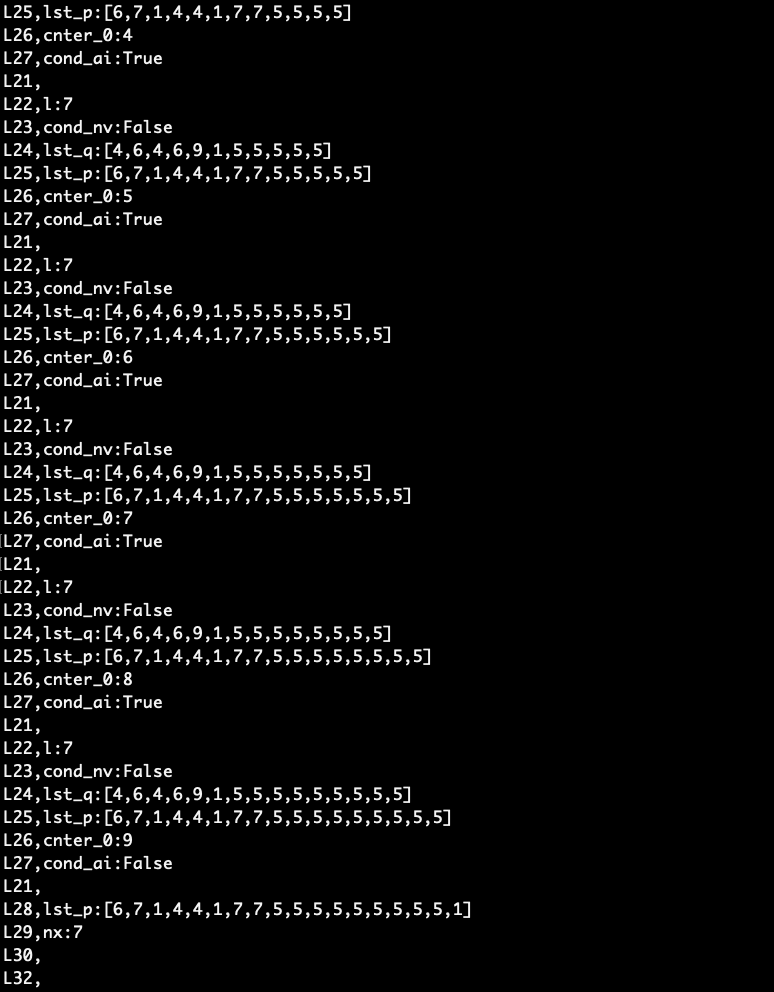}
    \label{fig:example-ptg}
\end{figure}

\paragraph{Program generation \& Trace format.} We follow L0-Bench to generate simple synthetic Python programs using generative grammar. Each program is a standalone Python function with coding segments varying among assignment, if block, while block and list operations (append/pop). We further reduce the difficulty of each line of code by enforcing the binary arithmetic operations to be only increment or decrement by 1; the scope depth is set to 1 to avoid nested loops; list indexing is also excluded. An example of PTG instance is shown in Figure~\ref{fig:example-ptg}. The traces are generated by executing the programs given sampled inputs and post-processing Python execution bytecodes. Specifically, each trace is of format \boxed{\texttt{line\_number,var\_name:var\_value}}. For code lines that do not have value updates, only \boxed{\texttt{line\_number,}} is required.

\paragraph{Data generation \& statistics.} We list the distribution of training and test data configs in Table~\ref{tab:data-config}. Each program is unique. Training and in-distribution (ID) test splits do not have any overlap. During training, with $p=0.5$, we shift the first line of the program (and correspondingly the trace) by an offset drawn between 1 and 100. Without doing so models fail to achieve non-zero accuracy for the \emph{longer programs} split due to unseen line number in the programs and traces. Preliminary experiments also show the importance of including examples of diverse program/trace lengths to avoid overfitting to certain trace steps; training data are generated by binning short (5-20 lines of code) programs and mid (21-40 lines of code) programs by their trace steps, details listed in Table~\ref{tab:train-data-sp}.

\paragraph{Task comparison \& Four factors.} The task program trace generation essentially forces models to perform reliable sequential operations like a computer~\citep{neuralgpu,lear-to-execute} with \emph{guaranteed correctness}. However, PTG differs from existing algorithmic tasks (such as string reversal, modular addition, permutation composition, or in general recognizing regular languages)~\citep{zhou2023algorithmstransformerslearnstudy,kazemnejad2023the,deletang2023neural,liu2023exposing,merrill2024the,allenzhu2025physicslanguagemodels1} in two key ways. First, PTG provides the algorithm \textit{explicitly} in the context, while existing tasks require models internalizing the fixed algorithms into parameters during training. In other words, the context window now serves as memory: the beginning part hosting the ``instruction memory'' and the trace/scratchpad/chain-of-thought~\citep{merrill2024cot} part `` data memory.''  Second, the inputs and algorithms are decoupled, i.e., programs now have additional complexity terms independent of input size. We further refine the setup by ~\citet{l0-bench} to encompass four factors for more targeted evaluation of program execution. Specifically, we focus on::
\begin{itemize}
    \item \textbf{Program length}: Given a fixed max window size (4096 in our experiments), the split \emph{longer programs} tests how models behave when the size of ``instruction memory'' increases. Longer programs also correspond to more complex procedures which often have higher branching factors. 
    \item \textbf{Trace steps}: Given a fixed max window size (4096 in our experiments), the split \emph{longer traces} tests how models behave when the size of ``data memory'' increases. This can be achieved by increasing the while block size and/or the while loop iteration numbers without increasing the program length. 
    \item \textbf{Number of variables}: The split \emph{more variables} tests whether models can handle programs with larger number of input variables, some of which may serve as distractors. 
    \item \textbf{Input size}: The split \emph{longer lists} tests if models can apply in-distribution algorithm to larger entity size. This is commonly studied in existing works, such as training models up to $n$-digit addition and testing on inputs with $[n+1,k]$ digit inputs. 
\end{itemize}

\paragraph{Scratchpad size.} Give a list of $T$ elements and a program of $N$ code lines with $k$ while loop (scope-depth $1$) iterations, PTG requires generation of $O(kNT)$ tokens, similar to executing a linear-time algorithm with explicit scratchpad generation. This contrasts with trace-free tasks that test whether models can perform equivalent computation entirely within hidden states.

\begin{table}[h]
    \centering
    \small
    \caption{Distribution of training and eval data configs. OOD dimensions are highlighted in blue. Program length in number of lines; Trace length in number of steps; input size in the length of the list. }
    \scalebox{0.85}{
    \begin{tabular}{lccccr}
    \toprule
    \textbf{Split} & \textbf{Program length} & \textbf{Trace steps} & \textbf{Num of variables} & \textbf{Input size} & \textbf{Avg. Out Tokens} \\
    \midrule
    \textsc{Train} & & & & & \\
    \hspace{1em} \textbf{Short program} & [5,20] & [5,103] & [1,10] & [5,10] &  -\\
    \hspace{1em} \textbf{Mid program} & [21,40] & [5,103] & [1,10] & [5,10] &  -\\
    \midrule
    \textsc{Eval} & & & & & \\
    \hspace{1em} \textbf{ID (short)} & [5,20] & [59,79] & [1,10] & [5,10] & 713.2  \\
    \hspace{1em} \textbf{ID (mid)} & (20,40] & [90,103] & [1,10] & [5,10] & 1052.3   \\
    \hspace{1em} \textbf{Longer lists} & (20,40] & [5,103] & [1,10] & \textcolor{blue}{[10,15]} & 828.1 \\
    \hspace{1em} \textbf{More variables} & (20,40] & [5,103] & \textcolor{blue}{[10,50]} & [5,10] & 644.8 \\
    \hspace{1em} \textbf{Longer traces - 1} & (20,40] & \textcolor{blue}{[121,142]} & [1,10] & [5,10] & 1580.5 \\
    \hspace{1em} \textbf{Longer traces - 2} & (20,40] & \textcolor{blue}{[172,300]} & [1,10] & [5,10] & 2402.3\\
    \hspace{1em} \textbf{Longer programs} & \textcolor{blue}{(40,60]} & [5,103] & [1,10] & [5,10] & 1279.3\\ \bottomrule
    \end{tabular}}
    
    \label{tab:data-config}
\end{table}

\begin{table}[h]
\small
\centering
\caption{Data split statistics. \textbf{pl}, program length in number of code liens; \textbf{tr} trace size in number of trace steps.}
\begin{tabular}{@{}lrr@{}}
\toprule
    \textbf{Train split}   & \textbf{Avg. Token per example} & \textbf{Total num. tokens} \\ \midrule
pl {[}5, 20{]} tr (5,7{]}      & 204.06   & 8.14E+07    \\
pl {[}5, 20{]} tr (7, 8{]}     & 208.19   & 8.31E+07    \\
pl {[}5, 20{]} tr (8, 10{]}    & 268.01   & 1.07E+08    \\
pl {[}5, 20{]} tr (10, 15{]}   & 341.37   & 1.36E+08    \\
pl {[}5, 20{]} tr (15, 40{]}   & 548.64   & 2.19E+08    \\
pl {[}5, 20{]} tr (40, 59{]}   & 749.28   & 2.99E+08    \\
pl {[}5, 20{]} tr (59, 79{]}   & 986.17   & 3.93E+08    \\
pl {[}20, 40{]} tr (14, 25{]}  & 615.42   & 2.46E+08    \\
pl {[}20, 40{]} tr (25, 48{]}  & 860.15   & 3.43E+08    \\
pl {[}20, 40{]} tr (48, 63{]}  & 1057.39  & 4.22E+08    \\
pl {[}20, 40{]} tr (63, 77{]}  & 1244.97  & 4.97E+08    \\
pl {[}20, 40{]} tr (77, 90{]}  & 1419.38  & 5.66E+08    \\
pl {[}20, 40{]} tr (90, 103{]} & 1600.17  & 6.38E+08    \\
       & \multicolumn{1}{r}{\textbf{Sum}}       & 4.03E+09   \\ \bottomrule
\end{tabular}
\label{tab:train-data-sp}
\end{table}

\section{Model description} \label{sec:mdl-desc}

\begin{table}[h]
    \small
    \caption{Model description}
    \begin{tabular}{p{0.32\linewidth}p{0.72\linewidth}}
    \toprule
    \textbf{Model} & \textbf{Description} \\
    \midrule
    \textbf{RoPE}~\citep{rope} & Rotate query/key vectors before softmax attention. \\
    \textbf{NoPE}~\citep{nope} & Not applying any explicit positional encoding. \\
    \textbf{SWAN}~\citep{swan} & Interweaving sliding window (RoPE) and global attention (NoPE) layers. \\
    \textbf{ALiBi}~\citep{alibi} & Attention logits added by $m_k(j-i)$ for $q_i$, $k_j$, and head-specific slope $m_k$. \\
    \textbf{NaPE}~\citep{nape} & Slope of latter half of ALiBi is set to 0. \\
    \textbf{Fox}~\citep{fox} & Data-dependent forget gate added to attention logits. \\
    \textbf{Canon}~\citep{canon} & Short 1-D depthwise causal convolution applied to position "abc." \\
    \textbf{STB}~\citep{stb} & Replace Softmax in attention with strong recency-biased stick-breaking. \\
    \textbf{$\alpha$-Entmax}~\citep{nape} & Sparse attention allowing 0 weight on tokens whose scores below a threshold. \\
    \textbf{PaTH}~\citep{path} &  Data-dependent positional encoding via cumulative Householder transformation.  \\
    \bottomrule
    \end{tabular}
    \label{tab:mdls}
\end{table}

\section{Experimental configurations} \label{sec:append-exp}

Detailed training configurations are provided in Table~\ref{tab:train-configs}. Transformer variants evaluated in this work are listed in Table~\ref{tab:mdls}. For STB and ASEntmax, we experiments with both RoPE and NoPE, and report the best for each (STB+RoPE, ASEntmax+NoPE). All models are trained with A100 GPUs. 

\begin{table}[h]
    \centering
    \small
    \caption{Training and model configuration summary.}
    \begin{tabular}{lr}
\toprule
\textbf{Config name} & \textbf{Value} \\
\midrule
\textbf{Number of layers} & 12 \\
\textbf{Number of heads} & 16 \\
\textbf{Embedding dim} & 1024 \\
\textbf{Feedforward} & SwiGLU \\
\textbf{Learning rate} & 3e-4 \\
\textbf{LR scheduler} & Cosine \\
\textbf{Adam Betas} & (0.95, 0.9) \\
\textbf{Weight decay} & 0.1 \\
\textbf{Warmup steps} & 1000 \\
\textbf{Total train steps} & 20000 \\
\textbf{Batch size} & 128 \\
\textbf{Sequence length} & 4096 \\
\textbf{Decoding} & Greedy \\
\midrule
\multicolumn{2}{c}{\textbf{Model-specific configurations}} \\
\midrule
\textbf{RoPE base} & 10000 \\
\textbf{ALiBi slopes} & $\text{slope}_i =(2^{-\frac{8}{n}})^{i+1}$ for $i = 0\dots n-1$\\
\textbf{SWAN layers} & [1,2,3,5,6,7,9,10,11] \\
\textbf{SWAN window size} & 512 \\
\textbf{Canon conv window} & 4 \\
\textbf{Canon added position} & "abc" \\
\textbf{ASEntmax $\alpha$} & 1.5 \\
\bottomrule
    \end{tabular}
    \label{tab:train-configs}
\end{table}

\begin{figure}[h]
    \centering
    \caption{NoPE + Canon accuracy across evaluation splits throughout training (average over 10 random seeds). In-distribution performance saturates much earlier than OOD splits. The learning speed for longer programs is much slower than other splits.}
    \includegraphics[width=1.0\linewidth]{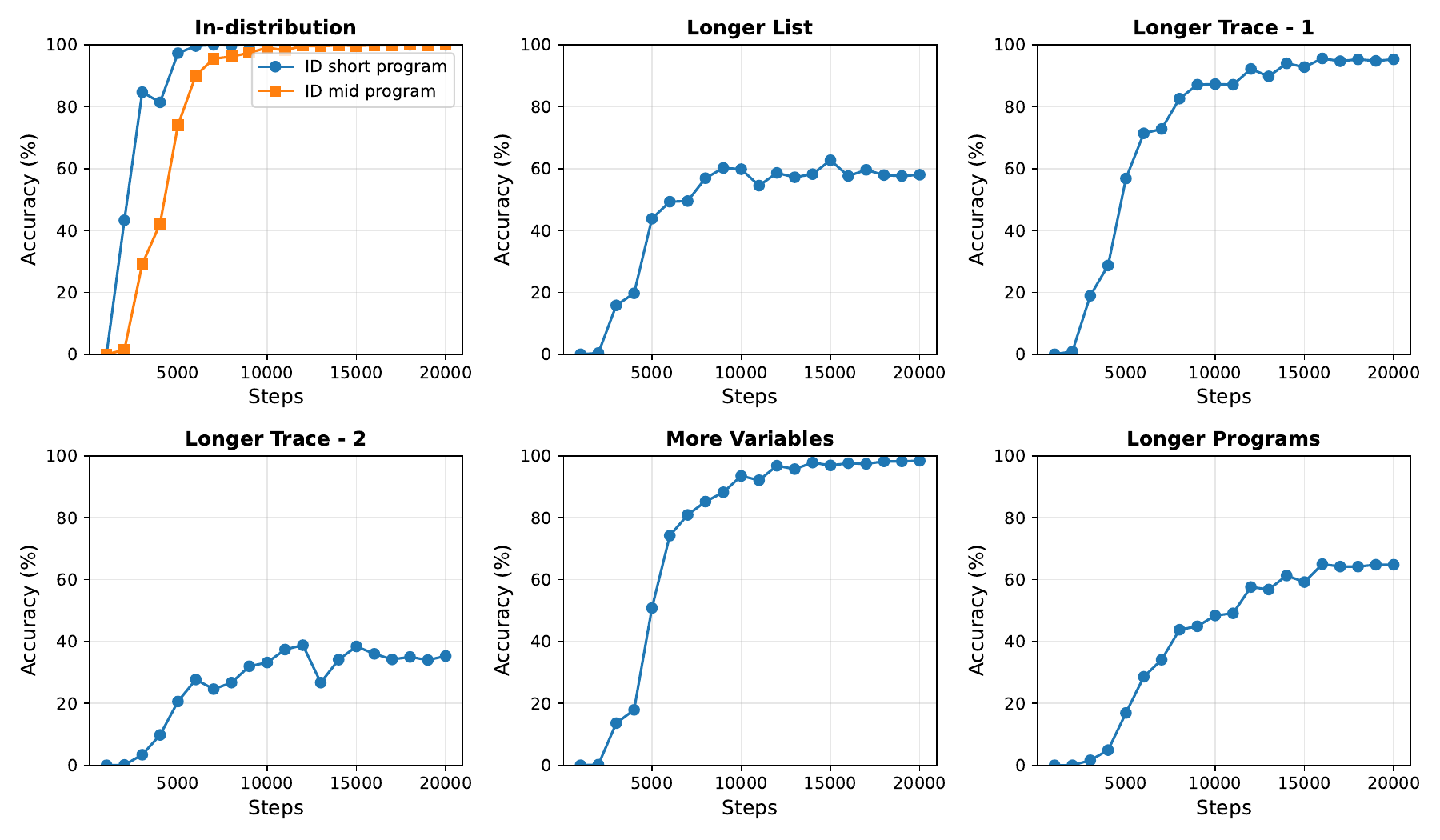}
    \label{fig:Accuracy across training.}
\end{figure}
\newpage

\end{document}